\documentclass[11pt,a4paper]{article}
\usepackage[hyperindex,breaklinks]{hyperref}
\usepackage[hyperref]{scil_2019_nps}
\usepackage{times}
\usepackage{linguex}
\usepackage{amssymb}
\usepackage{todonotes}
\usepackage{booktabs}

\usepackage{url}

\aclfinalcopy 

\title{Non-entailed subsequences as a challenge for natural language inference}

\author{R. Thomas McCoy \\
  Department of Cognitive Science \\
  Johns Hopkins University \\
  {\tt tom.mccoy@jhu.edu} \\\And
  Tal Linzen \\
  Department of Cognitive Science \\
  Johns Hopkins University \\
  {\tt tal.linzen@jhu.edu} \\}

\date{}

\begin{document}
\setlength{\Exlabelwidth}{0.5em}
\setlength{\SubExleftmargin}{1.5em}
\maketitle

\paragraph{Introduction:} Natural language inference (NLI) --- the task of determining whether a premise entails a hypothesis --- is a central challenge for natural language understanding systems \citep{condoravdi2003entailment, dagan2006pascal, bowman2015large}. The availability of large sets of premises and hypotheses generated through crowdsourcing has made it possible to train neural networks without explicit logical representations to perform this task; such systems have reached considerable accuracy on these data sets \citep{radford2018improving,kim2018semantic}. Recent studies have identified biases in these data sets which complicate the interpretation of these successes; for instance, 
statistical regularities
in crowdsourced hypotheses make it possible to reach substantial accuracy without even considering the premise \cite{gururangan2018annotation,poliak2018hypothesis}.
Since neural networks excel at capturing such statistical regularities, success on biased data sets may reflect fallible heuristics rather than deep language understanding, underscoring the need for a controlled experimental approach for evaluating NLI systems. To this end, we introduce a challenge set that targets the following possible heuristic:

\ex. \textbf{The subsequence heuristic:} Assume that a sentence entails all of its subsequences.

This heuristic is attractive to a statistical learner because it often yields the correct answer for NLI sentence pairs: 



\ex.\a. John likes Baltimore a lot. $\rightarrow$\\
John likes Baltimore.
\b. Roses are red, and violets are blue $\rightarrow$\\
Violets are blue.

The subsequence heuristic is not a generally valid inference strategy, however; for example, it incorrectly predicts that the following sentence pairs are instances of entailment: 

\ex. Alice believes Mary is lying. $\nrightarrow$\\
Alice believes Mary.\label{ex:nps}

\ex. The book on the table is blue. $\nrightarrow$\\
The table is blue.

\ex. The student sent the gift by Max yawned. $\nrightarrow$\\
The student sent the gift.\label{ex:passive}

We conjecture that pairs such as \ref{ex:nps}-\ref{ex:passive}, in which the hypothesis is a \textit{nonentailed, nonconstituent}
subsequence of the premise, are highly unlikely to be generated as potential contradictions by untrained annotators; consequently, they will not be available when training the model and will not be reflected in standard accuracy metrics.

We propose to create a challenge set that leverages the syntactic constructions illustrated in \ref{ex:nps}-\ref{ex:passive}, as well as other constructions, to generate sentence pairs in which the hypothesis is a nonentailed nonconstituent
subsequence of the premise. We demonstrate the viability of our approach with a set of sentences modeled after \ref{ex:nps}. These sentence are referred to in psycholinguistics as NP/S sentences (e.g., \citealt{pritchett1988garden}), because the verb (\textit{believe}) can take either a direct object noun phrase (NP) or a sentence (S) as its complement; the hypothesis \textit{Alice believes Mary} is the result of incorrectly assuming that the complement of the verb is the noun phrase \textit{Mary} instead of the sentence \textit{Mary is lying}.
We evaluate a number of competitive NLI models on this challenge set. To anticipate our results, the accuracy of these models was close to 0\% (when chance performance is 50\%), supporting the hypothesis that they rely on the subsequence heuristic.

\paragraph{Models:}

We assess the performance of five neural-network NLI models. All models consisted of bidirectional LSTMs trained in two stages, following \citet{wang2018glue}: first, on one of the pre-training tasks described below, and then on NLI (with a classifier predicting the labels \textit{entailment}, \textit{contradiction} and \textit{neutral}), using the MNLI data set \cite{williams2018multinli}. Our pre-training tasks were: NLI using the MNLI corpus, combinatory categorial grammar (CCG) supertagging using tags from CCGbank (derived from the Penn Treebank) \cite{hockenmaier2007ccgbank}, image generation from captions using the MS COCO data set \cite{lin2014microsoft}, and language modeling (LM) using the WikiText-103 corpus \citep{merity2016pointer}. We also tested a model without pre-training, in which the encoder had random weights but the classifier was still trained on MNLI. 

\paragraph{Data set creation:}

We generated premises using the template NP$_1$ V$_1$ S$_1$, where (i) NP$_1$ appeared as the subject of V$_1$ in the MNLI training corpus, (ii) the subject of S$_1$ appeared as the direct object of V$_1$ in the corpus, and (iii) S$_1$ appeared in the corpus (not necessarily as a complement of V$_1$). These conditions ensured that our examples were in the domain on which the models were trained, and that the models had been exposed to all words and dependencies in our examples. For example, based on the sentences in (6) from the MNLI training corpus, we generated the example in (7):

\ex.\a. The Knights believed that their goal was justified, however they would succumb to infighting.
\b. No one believed the story that Miss Howard has made up.
\c. San'doro said the story was awful.


\ex.The Knights believed the story was awful. $\nrightarrow$\\
The Knights believed the story.

We built our examples around the verbs \textit{heard}, \textit{believed}, \textit{felt}, and \textit{claimed}. We generated 200 sentence pairs and had each one annotated by three workers on Amazon Mechanical Turk. We kept the 88 examples for which two of the annotators agreed that the example made sense and that the correct label was \textit{not entailment}. Some premises from our data set shown are in \ref{ex:steelsphere}-\ref{ex:miners}, with the associated non-entailed hypotheses underlined:

\ex. \underline{They claimed the cinema} is in a steel sphere.\label{ex:steelsphere}


\ex. \underline{The committee felt the pressure} was applied by oversight entities.

\ex. \underline{They heard the miners} were prepared to fight.\label{ex:miners}


\paragraph{Results:}

Table \ref{tab:results} reports accuracies on the MNLI development set and our NP/S set.
All models performed reasonably well on MNLI but substantially below chance on the NP/S set. Closer inspection revealed that most examples that the models correctly labeled \textit{not entailment} had a negation word in the premise but not the hypothesis: 

\ex. They heard the tapes are of \textbf{no} importance $\nrightarrow$\\
They heard the tapes.

\ex. The young American believed the statistician is \textbf{not} involved.
 $\nrightarrow$\\
The young American believed the statistician.

This observation suggests that even when the models correctly labeled an NP/S example as \textit{not entailment} they may have done so using a heuristic that relied heavily on irrelevant negation words. To test whether this was the case, we removed all negation words from the NP/S examples; as shown in Table \ref{tab:results}, this caused the accuracy of all models to fall to nearly 0, suggesting that the models were indeed using a negation-word-based heuristic. 
Thus, even when the models provided the correct label on the NP/S evaluation set, they generally did so for the wrong reason.

\begin{table}
\centering
\begin{tabular} {lccc}
    \toprule
& MNLI & NP/S & NP/S (no neg.) \\ \midrule
MNLI & 0.75 & 0.08 & 0.01 \\
CCG & 0.67 & 0.17 & 0.03 \\
MSCOCO & 0.61 & 0.24 & 0.03 \\
LM & 0.72 & 0.06 & 0.00 \\
Random & 0.73 & 0.03 & 0.01 \\ \midrule
Chance & 0.33 & 0.50 & 0.50 \\ \bottomrule
\end{tabular}
    \caption{Accuracies on MNLI, our unmodified NP/S set, and our NP/S set with negation words removed.
} \label{tab:results}
\end{table}

\paragraph{Conclusions:}

All models perform poorly on the NP/S evaluation set, especially when irrelevant negation words are removed. These results indicate that standard neural models trained on crowdsourced NLI data sets
are prone to heuristics based on subsequences and negation and suggest that there is substantial room for improving the sophistication of NLI models. The clear and interpretable results of our evaluation strategy motivate expanding our data set to include additional constructions with similar properties, some of which are illustrated in \ref{ex:nps}-\ref{ex:passive}, to create an ambitious standard for measuring progress in NLI. In future work, we will also expand this data set into a more general test suite for evaluating which heuristics a model has learned. This test suite will include the subsequence heuristic and the negation heuristic from the current work, as well as other heuristics based on properties such as lexical overlap between the premise and the hypothesis.
We will also investigate other types of models trained on NLI, such as non-neural models  and tree-based neural models, to test whether reliance on the subsequence heuristic arises from the the NLI task or from the sequential nature of standard RNNs, or both.

\section*{Acknowledgments}

We thank the 2018 Jelinek Summer Workshop on Speech and Language Technology (JSALT) for support of this research, and the members of the JSALT General-Purpose Sentence Representation Learning team for comments during the project's development. 

This material is based upon work supported by the National Science Foundation Graduate Research Fellowship Program under Grant No. 1746891. Any opinions, findings, and conclusions or recommendations expressed in this material are those of the authors and do not necessarily reflect the views of the National Science Foundation.

\vspace{-0.4cm}
\bibliography{scil_2019_nps}
\bibliographystyle{acl_natbib}

\end{document}